\documentstyle[colap]{article}

\makeatletter
 \def\submitted#1{\setbox\@tempboxa\vbox{\normalsize \tt \raggedright
    #1 \\ \hbox{}}
    \vspace{-2.5 cm} \usebox\@tempboxa \\
    \vspace{-\ht\@tempboxa} \vspace{2.5 cm}}
\makeatother

\title{
\submitted{Appears in "Proceedings COLING96", Copenhagen, August 1996}
The discourse functions of Italian  subjects:\\
a centering approach}
\author{Barbara Di Eugenio \\
Computational Linguistics \\
Carnegie Mellon University \\
Pittsburgh, PA, 15213 USA \\
dieugeni@andrew.cmu.edu}
\date{}

\begin{document}
\thispagestyle{empty}

\maketitle

\begin{abstract}
This paper examines the discourse functions that different types of
subjects perform in Italian within the centering framework
\cite{gjw95}.  I build on my previous work \cite{coling90} that
accounted for the alternation of null and strong pronouns in subject
position.  I extend my previous analysis in several ways: for example,
I refine the notion of {\sc continue} and discuss the centering
functions of full NPs.
\end{abstract}

\vspace{0.25in}

\newcommand{\bdes}{\begin{description}}
\newcommand{\edes}{\end{description}}
\newcommand{\bite}{\begin{itemize}}
\newcommand{\eite}{\end{itemize}}
\newcommand{\benum}{\begin{enumerate}}
\newcommand{\eenum}{\end{enumerate}}
\newcommand{\bverb}{\begin{verbatim}}
\newcommand{\everb}{\end{verbatim}}
\newcommand{\m}[1]{{\em #1\/}}
\newcommand{\smc}[1]{\mbox{{\sc #1}}}
\newcommand{\rome}[1]{{\rm #1}}
\newcounter{thesubeg}
\newcounter{thesubegfoo}
\newenvironment{egs}[1]{\refstepcounter{equation}\label{#1}\samepage\begin{list}
{(\arabic{equation}\alph{thesubeg})}{\usecounter{thesubeg}}}{\end{list}}
\newenvironment{eg}[1]{\refstepcounter{equation}\label{#1}\samepage\begin{list}
{(\arabic{equation})}{\usecounter{thesubeg}}}{\end{list}}
\newcommand{\refegs}[2]{\setcounter{thesubegfoo}{#2}(\ref{#1}\alph{thesubegfoo})}
\newcommand{\refeg}[1]{(\ref{#1})}
\newcommand{\centc}{{\sc continue\/}}
\newcommand{\centcr}{{\sc ret-cont\/}}
\newcommand{\centr}{{\sc retain\/}}
\newcommand{\cents}{{\sc shift\/}}
\newcommand{\centss}{{\sc smooth-shift\/}}
\newcommand{\centrs}{{\sc rough-shift\/}}

\newcommand{\tref}[1]{Table~\ref{#1}}

\newtheorem{rul}{Rule}
\section{Introduction}

Interpreting referential expressions is important for any large
coverage NL system; while such systems do exist for Italian, e.g.
\cite{stock93,enzo94}, to my knowledge not much attention has been
devoted to the interpretation of Italian referential expressions.
Some exceptions are \cite{vieri90}, that discusses interpretation of
referential expressions within dialogues to access a videodisc on
Italian art; \cite{not95}, that adopts a systemic grammar approach
\cite{halliday76}; and \cite{coling90}, which uses centering theory
\cite{gjw95} to account for the alternation of null and
strong subjects. 

In this paper, I build on and expand \cite{coling90} in several ways.
First, I reanalyze the hypotheses I proposed earlier with respect to a
corpus of naturally occurring data:\footnote{The examples in
\cite{coling90} were constructed.} I show that those hypotheses are
basically supported; and that further insight can be gained
by looking at a two member sequence of centering
transitions rather than at just one transition.
Second, I extend
my previous analysis by also discussing the centering functions of full
NPs in subject position, and some occurrences of pronouns unaccounted
for by centering.

\section{Centering theory}

\label{cent}

Centering theory \cite{gjw86,lyn1,gjw95} models local
coherence in discourse: it keeps track of how local focus varies from
one utterance to the next. 
Centering postulates that:\footnote{The version of centering I
present here is  from \cite{lyn1}.}\\ 
$\bullet$ Each utterance U$_n$ has associated with it a set of discourse entities,
the {\sc forward-looking centers} or Cfs. 
The Cf list is ranked according to discourse salience.\\
$\bullet$ The {\sc backward-looking center}, or Cb, is the member of the Cf
list that U$_n$  most centrally concerns, and
that links U$_n$  to the previous discourse.\\
$\bullet$ Finally, the {\sc preferred center}, or Cp, is the highest ranked
member of the Cf list.  The Cp represents a prediction about the Cb of the
following utterance. 

Transitions between two adjacent utterances
U$_{n-1}$ and U$_{n}$ can be characterized as a function of \m{looking
backward} --- whether Cb(U$_n$) is the same as Cb(U$_{n-1}$)  --- and of
\m{looking forward} --- whether Cb(U$_n$) is the same as Cp(U$_{n}$).
Table~\ref{chart} illustrates the four transitions that are defined
according to these constraints.
\begin{table*}[htpb]
\begin{center}
\begin{tabular}{||c|c|c||}\hline \hline
& Cb(U$_n$) = Cb(U$_{n-1}$) & Cb(U$_n$) $\neq$ Cb(U$_{n-1}$)\\\hline \hline
Cb(U$_n$) = Cp(U$_{n}$) & {\sc continue} & {\sc smooth-shift}  \\ \hline
Cb(U$_n$) $\neq$ Cp(U$_n$) & {\sc retain} & {\sc rough-shift} \\ \hline \hline
\end{tabular}
\end{center}
\protect \caption{Centering Transitions}
\label{chart}
\end{table*}
\cite{lyn1} proposes a default ordering on transitions which
correlates with discourse coherence: \centc\/ is preferred to \centr\/
is preferred to \centss\/ is preferred to \centrs.\footnote{
\cite{gjw86,gjw95} propose that the ordering on
transitions pertains to {\em sequences} of transitions rather than
to single transitions.}

The saliency ordering on the Cf list, which is generally equated with
grammatical function, for Western languages is {\sc subject $>$
object2 $>$ object $>$ others}, where {\sc others} includes
prepositional phrases and adjuncts. \cite{kame} was the first to
point out that for languages such as Japanese  empathy
and topic marking affect the Cf ordering, and proposed the following
ranking 
\begin{eg}{cf-ranking2}
\item \ {\sc {\bf empathy} $>$  subject $>$ object2 $>$ \\ object $>$ others}
\end{eg}
I follow \cite{turan95} in adopting \refeg{cf-ranking2} also for
Western languages.  Turan argues that
a notion analogous to empathy arises in Western languages as well:
e.g.  with perception verbs, it is the
experiencer, which is often in object position, rather than the
grammatical subject, that should be ranked higher.

Finally, centering provides an interesting framework for studying the
functions of pronouns, as the observation that  the Cb is often deleted
or pronominalized can be stated as the following  rule:
\begin{rul} 
\label{pronoun-rule}
If some element of Cf(U$_{n-1}$) is realized as a pronoun in U$_n$,
then so is Cb(U$_n$).
\end{rul}

This rule has been computationally interpreted to individuate the Cb.
If U$_{n}$ has:
\begin{itemize}
\item  {\rm  a single pronoun, that is  Cb(U$_{n})$;}
\item  {\rm  zero or more than one pronoun, Cb(U$_{n})$ is:}
\begin{itemize}
\item {\rm Cb(U$_{n-1}$) if 
Cb(U$_{n-1}$) is realized in U$_{n}$;}
\item {\rm otherwise the highest ranked Cf(U$_{n-1}$) which
is realized in U$_{n}$.}
\end{itemize}
\end{itemize}

Let's apply centering to the constructed example in
\refeg{first}. 
In \refegs{first}{1} Cb~=~? because the
Cb of a segment initial utterance is left unspecified; in
\refegs{first}{2} the Cb is \m{John}, as it is the only pronoun, and
also the only entity belonging to the Cf list of \refegs{first}{1}
realized in \refegs{first}{2}.

\begin{egs}{first}
\item {\bf John} {\em is a nice guy.}\\
\begin{small}
{\tt Cb = ? Cf = [John]}
\end{small}
\item  {\bf He} {\em met {\bf Mary} yesterday.}\\
\begin{small}
{\tt Cb = John,   Cf = [John $>$  Mary]}
\end{small}
\item
\renewcommand{\theenumi}{\roman{enumi}}
\begin{enumerate}
\item {\bf He}  {\em likes}  {\bf her}. (\centc)\\
\begin{small}
{\tt Cb = John,  Cf = [John $>$  Mary]}
\end{small}
\item  {\bf She}  {\em likes } {\bf him}. (\centr) \\
\begin{small}
{\tt Cb = John, Cf = [Mary $>$ John]}
\end{small}
\item  {\bf She}  {\em was with} {\bf Lucy}. (\centss) \\
\begin{small}
{\tt Cb = Mary, Cf = [Mary $>$ Lucy]}
\end{small}
\item {\bf Lucy}  {\em was with} {\bf her}. (\centrs) \\
\begin{small}
{\tt Cb = Mary,  Cf = [Lucy $>$ Mary]} 
\end{small}
\end{enumerate}
\end{egs}

In \refegs{first}{3}.i we have a \centc, as its Cb is 
\m{John} (the highest entity on the Cf list of
\refegs{first}{2}), and so is its Cp.
In \refegs{first}{3}.ii, the Cb is still \m{John} as in
\refegs{first}{3}.i, but the Cp now is {\em Mary}, thus we have a \centr. In
both \refegs{first}{3}.iii and \refegs{first}{3}.iv the Cb is \m{Mary}
(the only entity belonging to the  Cf list in \refegs{first}{2} that is
realized): as \m{Mary} is also the Cp in \refegs{first}{3}.iii, 
a \centss\/ occurs. Instead, as \m{Lucy} is the
Cp in \refegs{first}{3}.iv, a \centrs\/ occurs.

Centering theory has appealing traits from both  cognitive and 
computational points of view. From a cognitive perspective, it
explains certain phenomena of local discourse coherence (e.g. pronominal
``garden paths''), and is  supported by psycholinguistic experiments
\cite{ggg93}.  Computationally, it is a simple mechanism, 
and thus  it has been the basis for  simple algorithms for 
anaphora resolution \cite{lyn1}.

Much work still remains to be done on centering.  For example, most
development so far has been based on simple constructed examples:
to apply centering to real text, issues such as how possessives and
subordinate clauses affect referring expression resolution must be
addressed. This paper is a contribution in that direction.  

\section{The Italian pronominal system}

\label{ital}

Italian has two pronominal systems \cite{cal}: weak pronouns, that must
always be cliticized to the verb (e.g. {\sf lo, le, gli} - respectively
him, accusative; them, feminine accusative or her, dative; him,
dative), and strong pronouns ({\sf lui, lei, loro} - respectively he
or him; she or her; they or them).\footnote{\label{egli}{\sf Lui,
lei, loro} are the oblique forms of the strong system, while the
nominative forms are respectively {\sf egli, ella, essi/e}: 
in current Italian the latter forms are   rarely used as the
oblique forms have replaced them in subject position --- 
in my corpus there are only four occurrences of these nominative
forms, and they all  occur  in the same article \cite{dick}.}  The
null subject is considered part of the system of weak pronouns.

Weak and strong pronouns are often in complementary distribution, as 
strong pronouns have to be used in prepositional phrases, e.g. {\sf
per lui}, \m{for him}. However, this
syntactic alternation doesn't apply in subject position. The choice of 
null versus strong pronoun depends on pragmatic factors; the centering
explanation offered in \cite{coling90} goes  as follows:
\begin{egs}{strategies2}
\item Typically, a null subject signals  a \centc, and a strong pronoun
a \centr\/ or a \cents.
\item A null subject can be felicitously used in cases of \centr\/ or
\cents\/ if in $U_{n}$ the syntactic context up to and including the
verbal form(s) carrying tense and / or agreement forces the null
subject to refer to a particular referent and not to Cb$(U_{n-1})$.
\end{egs}

The evidence for \refegs{strategies2}{2} provided in \cite{coling90} 
derived, among others, from modal and control verb constructions, in which
clitics may be cliticized to the infinitival complement of the higher
verb 
or may climb in front of the higher verb. When the clitic climbs,
certain pronominal ``garden path'' effects, deriving from a wrong
interpretation initially assigned to the null subject and later
retracted, are avoided.

\section{Italian subjects in discourse}

\label{meat}

\subsection{The corpus}

The corpus amounts to about  25 pages of text, and 12,000 words; it 
is composed of excerpts from two books
\cite{giardino,penelope}, a letter \cite{mila}, a posting on the 
Italian bulletin board \cite{bboard},
a short story \cite{cicala}, and three articles from two newspapers
\cite{metz,dick,perot}. The excerpts are of different lengths,
with the excerpts from the two books being the longest.

Texts were chosen to cover a variety of contemporary written Italian
prose, from formal (newspaper articles about politics and literature),
to informal (posting on the Italian bulletin board), and according to
the following criteria: a) minimal direct speech, which has
not been addressed in centering yet; b) prose that describes
situations involving several animate referents, because strong
pronouns can refer only to animate referents.

\tref{numbers3} shows the distribution of animate third person
subjects partitioned into: full NPs --- the numbers in parentheses
refer to possessive NPs; strong pronouns; null subjects --- I counted
only those whose antecedents are not determined by contraindexing
constraints \cite{chomsky81}.; other anaphors (e.g. {\sf tutte},
\m{all$_{fem}$}) --- they won't be analyzed in this paper.
\begin{table*}[htpb]
\small
\begin{center}
\begin{tabular}{||l||c||rr|ccc||}\hline \hline
Text & Total & \multicolumn{2}{c|}{Full NPs} & Strong & Zero & Other \\ \hline
\cite{giardino} & 111  &  45 & (11) &  23 & 36  & 7 \\[0cm]
\cite{penelope} & 17 &  6 & (0) & 2 & 9  & 0 \\
\cite{mila} & 8 &  1 & (0) & 2 & 4 &  1 \\
\cite{bboard} & 18  & 7 &(1) & 0 & 7  & 4 \\
\cite{cicala} & 40 &  26 & (1) & 1 & 13  & 0 \\
\cite{metz} & 36 & 28 & (6) & 1 & 6 & 1 \\
\cite{dick}  & 22 &  19 & (6) & 3 & 0 & 0 \\
\cite{perot} & 35 &  27 & (4) & 1 & 5 &  2 \\\hline
Total & 287 & 159 & (29) & 33 & 80  & 15 \\\hline \hline
\end{tabular}
\protect \caption{Animate 3rd person subjects}
\label{numbers3}
\end{center}
\end{table*}

\subsection{Issues}

When applying centering to real text, one realizes that many issues
have not been solved yet. I will comment here on how deictics,
possessives, and subordinate clauses affect centering.

\paragraph{Deictics} such as \m{I}, \m{you}, etc. The problem is whether they are 
part of the Cf list or not. I follow \cite{lyn-workshop} in
assuming that deictics are always available as part of global focus,
and therefore are outside centering.
\paragraph{Possessives.} 
\tref{dist-total} includes a category marked \m{poss\/}essive,
which refers to full NPs that include a possessive adjective referring
to an animate entity, such as \m{i suoi sforzi} --- \m{his efforts}.

The problem is how possessives affect Cb computation and Cf ordering.
While Cb computation does not appear to be affected by a possessive,
that behaves like a pronoun, the Cf ranking needs to be modified. An
NP of type \m{possess}ive refers to two entities, the possessor
P$_{or}$ and the possessed P$_{ed}$.  P$_{ed}$ corresponds to the full
NP, and thus its position in Cf is determined by the NP's grammatical
function; as regards P$_{or}$, my working heuristics is to rank it as
immediately preceding P$_{ed}$ if P$_{ed}$ is inanimate, as
immediately following P$_{ed}$ if P$_{ed}$ is animate. Such heuristics
appears to work, but needs to be rigorously tested.
\paragraph{Subordinates.}
Another important issue, that has not been extensively addressed yet
--- but see \cite{kame95,suri93} --- is how to deal with complex
sentences that include coordinates and subordinates. The questions
that arise concern whether there are independent Cb's and Cf lists for
every clause; if not, how  the Cb of the complex sentence is computed,
and how  semantic entities appearing in different clauses are ordered
on the global Cf list.  

In this paper, I will loosely adopt Kameyama's proposal
\shortcite{kame95} that sentences containing conjuncts and tensed
adjuncts are broken down into a linear sequence of centering
``units'', while tenseless adjuncts don't generate independent
centering units\footnote{The situation for complements is more
complicated, and space prevents me from discussing it.}.

\subsection{Centering Transitions}

\tref{dist-total} illustrates the distribution of referring expressions
with respect to  centering transitions.
\begin{table*}[htpb]
\small
\begin{center}
\begin{tabular}{||l||c||c|c|c||c||c||} \hline \hline
Type & Total &  {\sc continue} & {\sc 
retain} & {\sc shift} & {\sc Cent-est} & {\sc Other} \\ \hline \hline
zero & 80 & 56 & 4 & 6 & 12 & 2  \\ 
strong & 33 & 13 & 4 & 5 & 10 & 1 \\ \hline
NP & 81  & 17 & 11 & 7 & 44 & 2 \\
poss. & 25  & 11  & 5 & 1 & 8 & 0 \\\hline \hline
Total & 219 & 97 & 24 & 19 & 74 & 5 \\ \hline \hline
\end{tabular}
\end{center}
\protect \caption{Distribution of centering transitions}
\label{dist-total}
\end{table*}
\normalsize
The  number of full NPs in \tref{dist-total} is about half
their number in \tref{numbers3}: in fact,  full NPs
often introduce entities new to the discourse, in which case 
centering does not apply.

\tref{dist-total} includes two columns that don't refer to centering
transitions.  The column labeled {\sc cent-est} encodes referring
expressions that don't refer to a member of Cf(U$_{n-1}$), but to an
entity available in the discourse.  While such transitions do not
belong to centering, that models how centers change from one centering
unit to the next, they constitute referential usages of pronouns that
need to be explained. I call these transitions {\sc cent-estab}, for
{\sc center establishment}, because such references appear to
establish the new center of local discourse. Finally, {\sc other}
includes e.g. 
expressions that build a set out of Cb(U$_{n-1}$) and some other
entity, such as \m{sia lui che sua moglie} --- \m{both him and his
wife}.
It is not
clear how to deal with these constructions within the centering
framework, and thus, I have left them unanalyzed for the time being.

The results are as follows.  Null subjects are, not surprisingly, the
most frequently used expression --- 58\% --- for \centc's; the
difference between null subjects and all the other referring
expressions is also statistically significant ($\chi^2$~=~33.760,
p~$<$0.001).\footnote{$\chi^2$ test results are reported here more as
a source of suggestive evidence than as strong indicators, as the
observations in the corpus, which come from only 8 authors, are not
totally independent.} Vice versa, \centc's account for 70\% of null
subjects. However, even full NPs  can be used for \centc's.
As regards non possessive full NPs,  such
usages account for 16\% of
\centc's, and  for 20\% of  those NPs.
Also, 12\% of \centc's are encoded by means of possessive NP's, and
vice versa, 41\% of possessive NP's are used for \centc's.

The situation for \centr's and \cents's is not very clear, as none of
the four categories of referring expressions is predominant. All these
\cents's are actually \centss's, i.e., there are no
\centrs's at all. This is not surprising for null subjects, that are
never used for \centrs\/ \cite{turan95}, however it is puzzling for
full NPs. Apparently the Italian writers I selected adhere to the
default ranking of transitions, in which \centrs's are the least
preferred.

A significant difference in the usages of the four referring
expressions regards {\sc cent-est}. In this case, full NP's are used
70\% of the times, and the difference between full NP's, and all the
other expressions is significant ($\chi^2$~=~21.401, p~$<$0.001).

I will now  focus on the contrast between zeros and
strong pronouns,  in order to assess the strategies proposed in 
\refeg{strategies2}.

The first part of \refegs{strategies2}{1} --- 
null subjects used for \centc\/ --- 
is strongly supported. Zeros are used 80\% of the times, and there is
a significant difference ($\chi^2$~=~9.204, p~$<$~0.01) between zeros
and strong pronouns used in \centc\/ and zeros and strong pronouns
used in all other transitions taken together. Thus, in its use of null
subjects for \centc\/, Italian behaves in the same way as languages as
diverse as Japanese \cite{kame,lyn3,shima95} and Turkish
\cite{turan95}, (Turan, this volume).

However, as the percentage of strong pronouns used for \centc\/ is not
negligible, I set out to investigate which factors may affect such a
choice.  I analyzed the \centc's in my corpus, and I did find that one
relevant factor is the transition preceding the \centc\/ in question.
\tref{centcr} shows the different possible
transitions in U$_{n}$, that precedes U$_{n+1}$ in which a \centc\/
occurs.  The configuration in which a \centc\/ is preceded by a
\centr, which I call \centcr, differs from the other two because of
the constraint Cp(U$_n$)~$\neq$~Cb(U$_{n}$) in the \centr. This in a
sense predicts that the center will shift: but in a \centcr\/ such
prediction is not fulfilled. As \tref{centcr-dist} shows, this has
some consequences on the usage of null and strong pronouns.
\begin{table}[htpb]
\small
\begin{center}
\begin{tabular}{||l||c|c|c||}\hline \hline
& {\sc continue} & {\sc retain} & {\sc shift } \\ 
U$_n$&Cb$_n$=Cb$_{n-1}$&Cb$_n$=Cb$_{n-1}$&Cb$_n$$\neq$Cb$_{n-1}$
\\ &Cp$_n$=Cb$_{n}$&{\bf Cp$_n$$\neq$Cb$_{n}$}&Cp$_n$=Cb$_{n}$\\
\hline
U$_{n+1}$&\multicolumn{3}{c||}{Cb$_{n+1}$=Cb$_{n}$}\\
&\multicolumn{3}{c||}{Cp$_{n+1}$=Cb$_{n+1}$}\\ \hline \hline
\end{tabular}
\protect \caption{Transitions preceding a \centc}
\label{centcr}
\end{center}
\end{table}
\normalsize
\begin{table}[htpb]
\small
\begin{center}
\begin{tabular}{||l||c||c|c||} \hline \hline
Type & Total & {\sc cont-cont+} & {\sc ret-cont}  \\
&&{\sc shift-cont} & \\ \hline \hline
zero & 56 & 51 & 5   \\ 
strong & 13 & 7 & 6  \\ \hline \hline
Total & 69 & 58 & 11  \\ \hline \hline
\end{tabular}
\protect \caption{Pronoun occurrences for  \centcr}
\label{centcr-dist}
\end{center}
\end{table}
\normalsize
Compared to strong pronouns, null subjects are used 87\% of the times
for {\sc cont-cont} and {\sc shift-cont} taken together and only 45\%
of the times for \centcr.
Moreover, if a zero is used in a \centc, that \centc\/ is ten times
more likely to be a {\sc cont-cont} or {\sc shift-cont} than a
\centcr; in contrast, if a strong pronoun is used in a \centc, that
\centc\/ is as likely to be a {\sc cont-cont} or a {\sc shift-cont} as
a \centcr.\footnote{Also \cite{turan95} independently noticed the
existence of {\sc ret-cont}'s, and reports results similar to mine.}
These trends in usage are confirmed by  a strongly 
significant difference between zeros and strong pronouns used in {\sc
cont-cont} plus {\sc shift-cont}, and zeros and strong pronouns used
in \centcr\/ ($\chi^2$~=~10.910, p~$<$~0.001).
Fig.~\ref{irais} presents two examples of \centcr, one in
\refegs{ret-cont}{3} realized with a strong pronoun, the second in
\refegs{ret-cont}{5} realized with a null subject. In the utterance 
preceding \refegs{ret-cont}{1}, Cb~=~Irais and Cf~=~[Irais].
\begin{figure}[htb]
\begin{egs}{ret-cont}
\item {\em $\Phi$ Incomincer\'{o} a ricondurre 
il {\bf suo pensiero} sui {\bf suoi doveri} chiedendo\/{\bf le} ogni giorno}\\
\small
(I) will start to bring  {\bf her thoughts} back to {\bf her duties} by  
asking {\bf her} every day \\
{\tt Cf:[Irais $>$ I's thoughts, I's duties], 
Cb:Irais, continue}
\item \normalsize
{\em come sta {\bf suo marito}}.\\
\small
how {\bf her husband} is.\\
{\tt Cf:[husband $>$ Irais], Cb:Irais, retain}
\normalsize
\item {\em Non \`{e} che {\bf lei} {\bf gli} voglia granch\'{e} bene,} \\
\small
It's not the case that {\bf she} cares much about {\bf him}\\
{\tt Cf:[Irais $>$ husband], Cb:Irais, continue}
\normalsize
\item {\em perch\'{e} {\bf lui} non corre ad aprir\/{\bf le} la porta \\}
\small
because {\bf he} doesn't run to open the door {\bf for her}\\
{\tt Cf:[husband $>$ Irais], Cb:Irais, retain}
\normalsize
\item {\em ogni volta che $\Phi$ si alza per lasciare la stanza;}\\
\small
whenever {\bf (she)} gets up to leave the room.\\
{\tt Cf:[Irais], Cb:Irais, continue}
\normalsize
\end{egs}
\caption{Examples of \centcr}
\label{irais}
\end{figure}

As far as \centr's and \cents's go, the numbers are both too small to
draw any conclusion, and they don't seem to identify any preferred
usage for strong pronouns, contrary to what claimed by
\refegs{strategies2}{1}; also in the case of {\sc cent-est} there
doesn't seem to be any significant difference in usage. A topic for
future work is to verify whether there are any factors affecting the
choice between null and strong pronouns in these cases, especially
because null subjects used for \cents\/ or for {\sc cent-est}
sometimes result in a slightly less coherent discourse.

The second part of the claim,
\refegs{strategies2}{2} --- a null subject can be used if U$_{n}$
provides syntactic clues that force the null subject not to refer to
Cb(U$_{n-1}$) --- is supported; however, given the small numbers (four
\centr's and six {\cents}'s) this conclusion can just be
tentative. The most frequent clue is agreement in gender and / or
number.

\section{Conclusions}

In this paper, I examined the referring functions that different
types of subjects perform in Italian within the centering framework.
I built on the analysis presented in \cite{coling90}, and extended it
in several directions: first, I  used a corpus of really occurring
examples; second, I  included phenomena such as possessives and
subordinate clauses; third, I  refined the notion of \centc\/ by
pointing out the peculiarity of \centcr's; fourth, I  included
full NPs; fifth, I illustrated a type of pronominal usage, {\sc cent-est},
outside the purview  of centering.

Future work includes further analysis of  a somewhat surprising 
finding from the current study, i.e.   that NP's
encoding \centc's are not so rare. It is worth while to examine the
data further, to see under which conditions a full NP is licensed to
encode a \centc.  I also want to collect more \centcr's, \centr's, and 
\centss's to refine the analysis presented in this paper. 
Finally, another topic of research is {\sc cent-est}, even if it
is outside the centering framework, and under what conditions zeros
are used to encode it.

\end{document}